%% file: main.tex
\documentclass[letterpaper]{article}
\usepackage[preprint]{aaai2027}
\usepackage[hyphens]{url}
\usepackage{graphicx}
\usepackage{natbib}
\usepackage{caption}
\usepackage{algorithm}
\usepackage{algorithmic}

\usepackage{newfloat}
\usepackage{listings}
\DeclareCaptionStyle{ruled}{labelfont=normalfont,labelsep=colon,strut=off}
\floatstyle{ruled}
\newfloat{listing}{tb}{lst}{}
\floatname{listing}{Listing}

\usepackage{booktabs}
\usepackage{colortbl}
\usepackage{amsmath}
\usepackage{amssymb}

\title{RP-OPSD: Resolution-Privileged On-Policy Self-Distillation for Multimodal Large Language Models}
\author{
Qihui Zhu\textsuperscript{\rm 1}\equalcontrib,
Yuchen Wang\textsuperscript{\rm 1}\equalcontrib,
Zijian Wen\textsuperscript{\rm 1},
Tao Zhang\textsuperscript{\rm 1},
Mengjie Zhang\textsuperscript{\rm 1},
Yang Liu\textsuperscript{\rm 2},\\
Shuangwu Chen\textsuperscript{\rm 1}\corresponding,
Siying Wu\textsuperscript{\rm 1},
Jian Yang\textsuperscript{\rm 1},
Xiaofeng Jiang\textsuperscript{\rm 1}
}
\affiliations{
\textsuperscript{\rm 1}University of Science and Technology of China\\
\textsuperscript{\rm 2}ChangXin Memory Technologies, Inc.
}

\begin{document}

\maketitle

\begin{abstract}
On-Policy Self-Distillation (OPSD) uses privileged information available only to the teacher to provide dense token-level supervision on trajectories generated by the student. However, existing methods often rely on verified solution traces, explanations generated by external models, or manually localized visual evidence, which limits their scalable application to multimodal large language models. To address this issue, we exploit the information gap between high- and low-resolution views of the same image and propose RP-OPSD (Resolution-Privileged On-Policy Self-Distillation for Multimodal Large
Language Models). During training, the student policy generates on-policy trajectories from images at one-quarter of the original resolution, while the teacher policy provides supervision using the original-resolution images. By minimizing the divergence between their output distributions along the student trajectories, the student learns the predictive behavior of the teacher under high-resolution inputs, thereby strengthening its
low-resolution capability and transferring the learned improvement to original-resolution inference. RP-OPSD requires neither additional human annotations nor external models to generate solution traces but only image--question pairs. Experiments on Qwen3.5-9B show that RP-OPSD achieves a 5.45\% relative improvement in average performance at the original resolution and a $1.78\times$ training speedup over OPSD. These results demonstrate that resolution differences can serve as a simple and scalable source of privileged information, providing an effective and efficient approach to on-policy self-distillation for multimodal large language models.
\end{abstract}

\begin{links}
    \link{Code}{https://github.com/sansanyuchen/RP-OPSD}
\end{links}

\section{Introduction}
Recent advances in multimodal large language models (MLLMs) have substantially improved visual question answering, document understanding, chart analysis, and multimodal problem solving~\cite{liu2023llava,wang2024qwen2vl}. Recent open-weight systems such as Qwen3-VL~\cite{bai2025qwen3vl} and InternVL3.5~\cite{wang2025internvl35} further strengthen multimodal reasoning and efficient visual processing. However, efficient post-training remains challenging. Existing approaches mainly include supervised fine-tuning, reinforcement learning with verifiable rewards, and knowledge distillation~\cite{hinton2015distilling,shao2024deepseekmath}. Supervised fine-tuning learns from high-quality demonstrations, but its fixed training trajectories may differ from those generated at inference time, causing distribution shift and exposure bias~\cite{agarwal2024gkd}. Reinforcement learning uses model-generated trajectories, but sparse sequence-level rewards provide limited token-level feedback, while sampling multiple responses per prompt is costly~\cite{shao2024deepseekmath,huang2025visionr1,shen2025vlmr1}. Knowledge distillation provides dense supervision from a teacher's output distribution, but its reliance on offline trajectories leads to a similar train--inference mismatch~\cite{agarwal2024gkd}.

\begin{figure}[t]
    \centering
    \includegraphics[width=\columnwidth]
    {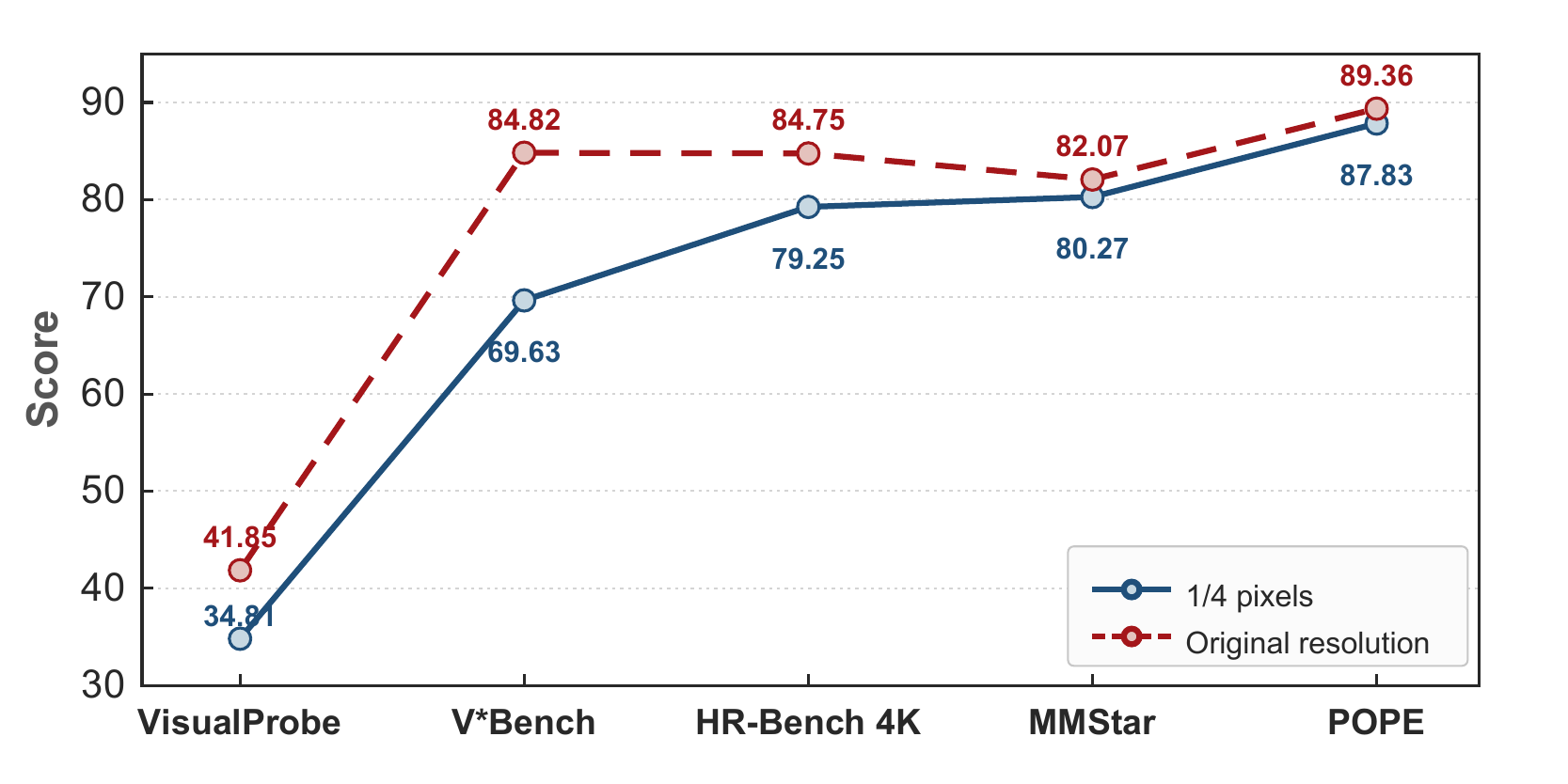}
    \caption{
        Resolution-induced capability gap of Qwen3.5-9B Base across five multimodal benchmarks. Reducing both image width and height by half results in an average performance drop of 6.21 points.
    }
    \label{fig:resolution_capability_gap}
\end{figure}

On-Policy Distillation (OPD) allows the student policy to generate its own trajectories, while the teacher policy provides token-level supervision at the states visited by the student~\cite{agarwal2024gkd,lu2025onpolicydistillation}. By matching their output distributions along these trajectories, OPD combines the advantages of on-policy learning and dense supervision. However, conventional OPD typically relies on a separate and stronger teacher model and requires compatible output spaces or model architectures, limiting its use in large-scale model post-training.On-Policy Self-Distillation (OPSD) further removes the need for an external teacher model~\cite{zhao2026opsd,shenfeld2026self,hubotter2026reinforcement}. It uses the same model as both the student and teacher under different input conditions. The student generates trajectories without additional information, while the teacher uses verified solutions, expert demonstrations, or rich environment feedback as training-time privileged information to provide token-level supervision. In this way, OPSD combines on-policy trajectories, dense feedback, and self-supervision, allowing the model to improve using its existing capabilities.

Although OPSD has shown promise in language tasks requiring multi-step reasoning~\cite{zhao2026opsd,shenfeld2026self,hubotter2026reinforcement}, extending it to multimodal large language models remains challenging because effective privileged information must be constructed for the teacher. Existing methods typically use reference answers, verified solution traces, or in-context examples, which are well suited to text tasks with clearly defined answers. However, errors in multimodal models may arise from missed objects, lost visual details, inaccurate localization, or weak cross-modal connections, often requiring additional region annotations, multimodal explanations, or evidence localization. OmniOPSD uses external models to generate multimodal solution evidence~\cite{cheng2026omniopsd}, while Vision-OPD constructs local privileged views by cropping relevant image regions for the teacher~\cite{visionopd}. Although these methods demonstrate the feasibility of teacher--student self-distillation under different visual conditions, they rely on external generation, object recognition, region segmentation, or local cropping, increasing the cost of data construction and quality control. Moreover, additional context does not always provide an effective teacher signal, as the performance of OPD can also depend on teacher selection, student capability, and the supervision context~\cite{mopd}. Therefore, constructing simple and effective privileged information remains a key challenge in extending OPSD to multimodal large language models.

These limitations motivate us to seek a simpler source of privileged information with three desired properties: it should create a meaningful capability gap between the teacher and student, preserve the semantic content of their inputs, and require neither external models nor additional annotations. Image resolution naturally satisfies these requirements. Given the same image and question, a multimodal large language model may produce a correct answer from the original-resolution image but an incorrect answer from its downsampled version. We refer to this phenomenon as the resolution-induced capability gap.As shown in Figure~\ref{fig:resolution_capability_gap}, when both the width and height of the input image are reduced to one-half of their original sizes, the average performance of Qwen3.5-9B Base~\cite{qwen2026qwen35} across five benchmarks decreases by 6.21 points. On V$^{*}$Bench~\cite{wu2023vstar}, the performance drop reaches 15.19 points. This gap arises because original-resolution images preserve more complete information about local textures, text, and small objects, whereas downsampling weakens these visual cues to different degrees.We therefore propose using an original-resolution teacher to supervise trajectories generated by a low-resolution student. This supervision first improves the model's capability under limited visual evidence. Because the teacher and student share the same model parameters, the resulting improvement can further transfer to original-resolution inference.

Building on this observation and hypothesis, we propose \textbf{RP-OPSD}, a resolution-privileged on-policy self-distillation framework for multimodal large language models. Given an original image, we instantiate two policies from the same multimodal model under different visual conditions. The student policy takes a low-resolution image and its corresponding question as input and samples on-policy responses from the current policy. The teacher policy instead uses the original-resolution image as a privileged visual view and provides supervision on the trajectories generated by the student. Rather than treating the resolution-induced capability gap only as a low-resolution robustness problem, RP-OPSD uses this gap to construct the asymmetric teacher--student conditions required by OPSD.Because this asymmetry is directly created from two resolutions of the same input, the optimization \textbf{requires only image--query pairs} and does not rely on annotated answers, externally generated reasoning traces, or localized visual evidence. This design substantially reduces the cost of constructing privileged information and can be readily applied to existing multimodal data.We evaluate RP-OPSD on Qwen3.5-4B and Qwen3.5-9B across a range of widely used multimodal benchmarks. Under half-resolution evaluation, RP-OPSD improves the average performance of Qwen3.5-9B by 6.09 points. Compared with representative post-training methods and state-of-the-art multimodal OPSD baselines, RP-OPSD achieves the largest average performance gains under original-resolution settings, with average relative improvements of 6.28\% and 5.45\% on Qwen3.5-4B and Qwen3.5-9B, respectively. These consistent improvements across resolutions jointly support our hypothesis. In addition, efficiency analysis on Qwen3.5-9B shows that RP-OPSD achieves a $1.78\times$ training speedup over OPSD.

In summary, our main contributions are as follows:
\begin{itemize}
    \item We propose \textbf{RP-OPSD}, a resolution-privileged on-policy self-distillation framework in which an original-resolution teacher supervises a low-resolution student on the student's rollouts through token-level distribution matching.

    \item RP-OPSD constructs privileged supervision directly from image--query pairs, without additional annotations or external models, enabling simple and scalable multimodal OPSD.

    \item Comprehensive experiments across two model scales and multiple multimodal benchmarks validate the effectiveness and efficiency of RP-OPSD.
\end{itemize}
\begin{figure*}[t]
    \centering
    \includegraphics[width=\linewidth]
    {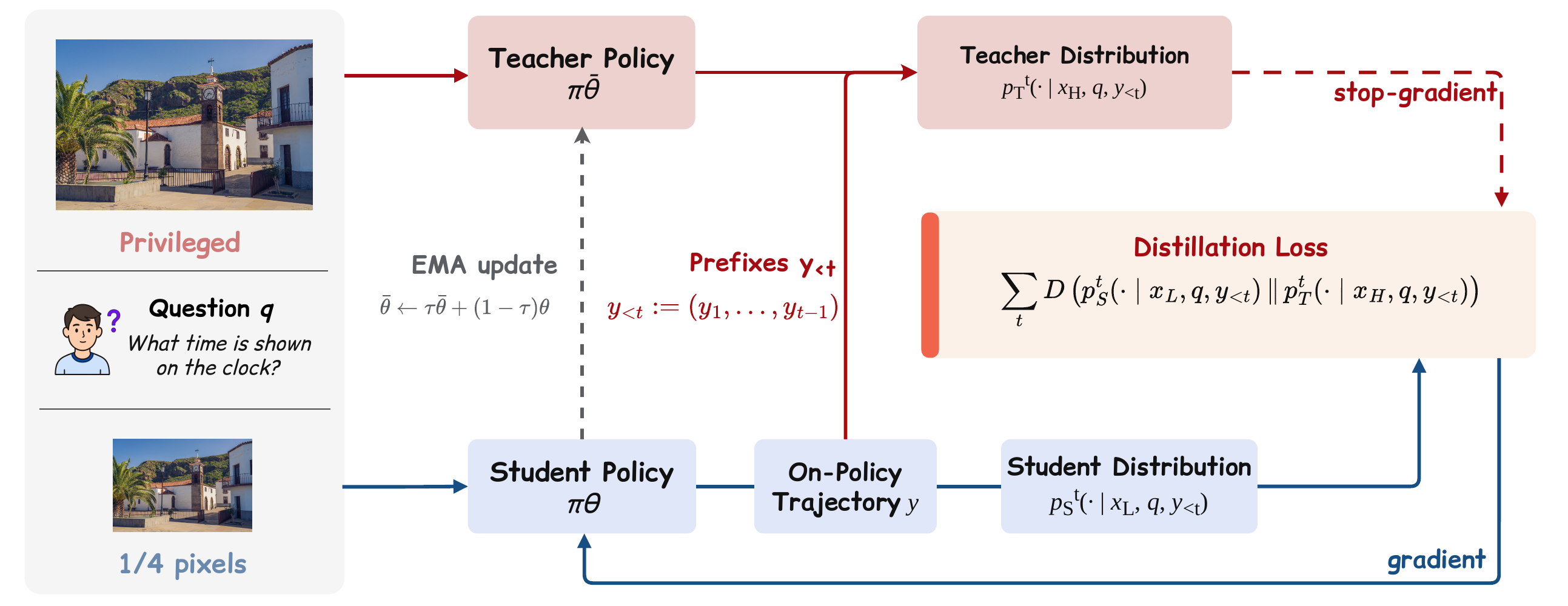}
    \caption{
    Overview of RP-OPSD. We first use a low-resolution student to generate
    on-policy trajectories from downsampled images. Then, an
    original-resolution teacher evaluates the same generated prefixes and
    provides token-level distribution targets with richer visual evidence.
    We optimize the student using the bias-corrected teacher-top-$K$ reverse
    KL objective and update the teacher with an exponential moving average,
    enabling self-distillation without external teachers or additional
    annotations.
    }
    \label{fig:overview}
\end{figure*}
\section{Related Works}
\subsection{Multimodal Large Language Models}

Multimodal large language models (MLLMs) typically map visual representations into the semantic space of large language models through vision encoders and modality connectors, and acquire visual understanding and cross-modal reasoning capabilities through multimodal pretraining and instruction tuning. Recent representative models, including Qwen3-VL~\cite{bai2025qwen3vl}, InternVL3.5~\cite{wang2025internvl35}, and GLM-4.5V~\cite{vteam2026glm45vglm41vthinkingversatilemultimodal}, have made significant progress in visual question answering, document understanding, fine-grained visual perception, and complex multimodal reasoning through large-scale and diverse multimodal pretraining, together with continued advances in vision--language alignment, cross-modal feature fusion, dynamic-resolution processing, and multi-stage post-training. However, their performance remains sensitive to input image quality and resolution. Reducing image resolution can weaken fine-grained visual cues, such as text, small objects, and local textures, leading to poorer understanding and reasoning on the same image--question pair. This sensitivity suggests that different input resolutions not only affect model performance but may also create a natural capability gap within the same model, providing a new way to establish teacher--student conditions for multimodal self-distillation.

\subsection{On-Policy Distillation}

Conventional knowledge distillation aligns the output distributions of the teacher and student on fixed trajectories~\cite{hinton2015distilling}, which may cause a state-distribution mismatch between training and inference. On-Policy Distillation (OPD) addresses this issue by allowing the student to generate its own trajectories while the teacher provides token-level supervision on the states actually visited by the student~\cite{agarwal2024gkd,lu2025onpolicydistillation}. OPD combines on-policy sampling with dense feedback, but it typically relies on a separate and stronger teacher model. On-Policy Self-Distillation (OPSD) further instantiates the same model as the teacher and student under different contexts~\cite{zhao2026opsd,shenfeld2026self,hubotter2026reinforcement}. The teacher uses verified solutions, expert demonstrations, or rich environment feedback as privileged information to supervise student-generated trajectories, removing the need for an external teacher model.

In multimodal settings, OmniOPSD uses multimodal explanations generated by an external model as teacher-side privileged information~\cite{cheng2026omniopsd}, while Vision-OPD uses cropped evidence regions to supervise a student conditioned on the full image~\cite{visionopd}. These methods require additional explanation generation or region construction. In contrast, RP-OPSD directly uses the original-resolution and downsampled views of the same image to create asymmetric teacher--student conditions. It requires no external generation, region annotation, or question synthesis, providing a simpler and more scalable way to construct privileged information.

\section{Method}
\label{sec:method}

We propose \textbf{RP-OPSD}, a resolution-privileged on-policy
self-distillation framework for multimodal large language models (MLLMs).
As shown in Fig.\ref{fig:overview}, a low-resolution student first generates
on-policy trajectories. A high-resolution teacher then provides token-level
distribution supervision over the same generated prefixes, and the student is
optimized by matching the teacher distribution on a teacher-selected support.

\subsection{Resolution-Privileged Formulation}
\label{sec:formulation}

Given an original-resolution image $x^H$ and a visual question $q$, we construct
a low-resolution view as
\begin{equation}
    x^L = \mathcal{R}_{1/2}(x^H),
    \label{eq:downsample}
\end{equation}
where $\mathcal{R}_{1/2}$ reduces both the image width and height by half.
Therefore, $x^L$ contains approximately one quarter of the original pixels.
The two views contain the same scene and field of view, without cropping,
region annotations, or additional location prompts. This aligned setting
allows us to use the visual details available in $x^H$ as privileged
information for learning from $x^L$.

RP-OPSD uses a low-resolution student and a high-resolution teacher with the
same model architecture:
\begin{equation}
    \pi_\theta^L(\cdot\mid x^L,q),
    \qquad
    \pi_\phi^H(\cdot\mid x^H,q).
    \label{eq:student-teacher}
\end{equation}
Both models are initialized from the same pretrained checkpoint,
$\theta_0=\phi_0$. Thus, the teacher's advantage comes from its access to
higher-resolution visual evidence rather than a larger model. After training,
the EMA teacher and rollout copy are discarded, and the optimized model can be
evaluated with either resolution without an additional teacher branch.

\subsection{Resolution-Privileged On-Policy Self-Distillation}
\label{sec:on-policy-distillation}

Distillation on fixed or teacher-generated responses may supervise the student
on prefixes that it rarely visits at inference time. Following the on-policy
distillation principle~\cite{agarwal2024gkd,lu2025onpolicydistillation,visionopd}, RP-OPSD instead performs distillation
on trajectories sampled from the current low-resolution policy. For each input
$(x^L,x^H,q)$, the rollout policy generates $G$ responses:
\begin{equation}
    y^{(g)}
    \sim
    \pi_{\theta^-}^{L}(\cdot\mid x^L,q),
    \qquad
    g=1,\ldots,G,
    \label{eq:student-rollout}
\end{equation}
where $\theta^-$ denotes the rollout policy synchronized with the student
before each rollout batch. Each batch is used for a single student update
before the next synchronization.

At token position $t$, the student and teacher evaluate the same
student-generated prefix:
\begin{equation}
    p_{g,t}(v)
    =
    \pi_\theta^L
    \left(
        v\mid x^L,q,y_{<t}^{(g)}
    \right),
    \label{eq:student-distribution}
\end{equation}
\begin{equation}
    r_{g,t}(v)
    =
    \operatorname{sg}
    \left[
        \pi_\phi^H
        \left(
            v\mid x^H,q,y_{<t}^{(g)}
        \right)
    \right],
    \label{eq:teacher-distribution}
\end{equation}
where $v\in\mathcal{V}$ is a vocabulary token and
$\operatorname{sg}[\cdot]$ stops gradients through the teacher. The teacher
does not generate a separate target response. Instead, it provides the
next-token distribution at states visited by the student. Since both models
receive the same question and text prefix, their predictions differ mainly
because of the available visual resolution. This design transfers
high-resolution visual knowledge while keeping the training states aligned
with the student's current behavior.

We maintain the teacher as an exponential moving average (EMA) of the student,
following the weight-averaged teacher paradigm~\cite{tarvainen2017meanteacher}.
After each successful student update, the teacher parameters are updated by
\begin{equation}
    \phi_{s+1}
    =
    (1-\rho)\phi_s+\rho\theta_{s+1},
    \label{eq:ema-update}
\end{equation}
where $s$ is the optimization step and $\rho$ is the EMA update rate. The
teacher is always evaluated without gradients. EMA provides a slowly changing
training target while allowing the teacher to follow improvements in the
student.

\subsection{Bias-Corrected Teacher-Top-$K$ Reverse KL Distillation}
\label{sec:topk-distillation}

Matching full-vocabulary distributions at every response token is expensive.
We therefore restrict the comparison to the tokens preferred by the
high-resolution teacher. At each token position, we define the teacher-selected
support as
\begin{equation}
    \mathcal{S}_{g,t}^{K}
    =
    \operatorname{TopK}_{v\in\mathcal{V}}
    r_{g,t}(v).
    \label{eq:teacher-topk}
\end{equation}
We gather the student and teacher probabilities on the same token indices.
These values remain probabilities from their original full-vocabulary softmax
distributions: we do not renormalize them within
$\mathcal{S}_{g,t}^{K}$ or add a separate tail bucket. Selecting the support
from the teacher ensures that the comparison covers the tokens most strongly
supported by the privileged visual input.

Naively restricting reverse KL to the teacher-selected top-$K$ support
introduces truncation bias because the retained probability masses do not sum
to one. Following the top-$k$ distillation objective in MOPD~\cite{mopd}, we
define the bias-corrected teacher-top-$K$ reverse KL as
\begin{equation}
\begin{aligned}
    d_K(p_{g,t}\|r_{g,t})
    =
    \sum_{v\in\mathcal{S}_{g,t}^{K}}
    \Bigg[
        &p_{g,t}(v)
        \log\frac{p_{g,t}(v)}{r_{g,t}(v)}
    \\[-2pt]
        &-p_{g,t}(v)+r_{g,t}(v)
    \Bigg].
\end{aligned}
\label{eq:corrected-rkl}
\end{equation}
The correction terms $-p_{g,t}(v)+r_{g,t}(v)$ correct this top-$K$ truncation
bias, ensuring that the gradient becomes zero when the student matches the
teacher on the retained support. This preserves the desired optimization
behavior without comparing the full vocabulary.

Under this synchronized one-update setting, we do not apply additional
rollout-policy importance reweighting. Let $m_{g,t}\in\{0,1\}$ denote the
response mask, which excludes prompt and padding tokens. The final RP-OPSD
objective is
\begin{equation}
    \mathcal{L}_{\mathrm{RP\text{-}OPSD}}
    =
    \mathbb{E}
    \left[
        \frac{
            \displaystyle
            \sum_{g,t}
            m_{g,t}
            d_K\!\left(p_{g,t}\|r_{g,t}\right)
        }{
            \displaystyle
            \sum_{g,t}m_{g,t}
        }
    \right].
    \label{eq:rp-opsd-objective}
\end{equation}
The expectation is taken over training examples and responses sampled from the
rollout policy. This objective provides dense distribution-level supervision
over the states visited by the low-resolution student.

RP-OPSD does not use answer-level rewards, supervised target responses, or a
separate reference policy. The multiple rollouts are used to cover diverse
student states rather than to perform group-based reward optimization.

Algorithm~\ref{alg:rp-opsd} summarizes the complete training procedure of
RP-OPSD.

\begin{algorithm}[tb]
\caption{Training procedure of RP-OPSD.}
\label{alg:rp-opsd}
\begin{algorithmic}[1]
\REQUIRE Training set $\mathcal{D}=\{(x_i^H,q_i)\}$; pretrained parameters
$\theta_0$; number of rollouts $G$; support size $K$; EMA rate $\rho$
\ENSURE Trained student policy $\pi_\theta^L$
\STATE Initialize student, teacher, and rollout parameters:
$\theta\leftarrow\theta_0$, $\phi\leftarrow\theta_0$,
$\theta^-\leftarrow\theta_0$
\FOR{each minibatch $\mathcal{B}\subset\mathcal{D}$}
    \FOR{each $(x^H,q)\in\mathcal{B}$}
        \STATE Construct the low-resolution view
        $x^L\leftarrow\mathcal{R}_{1/2}(x^H)$
        \STATE Sample $G$ responses
        $y^{(g)}\sim\pi_{\theta^-}^{L}(\cdot\mid x^L,q)$
        \FOR{each response $y^{(g)}$ and token position $t$}
            \STATE Compute the student distribution
            $p_{g,t}$ using $(x^L,q,y_{<t}^{(g)})$
            \STATE Compute the stopped-gradient teacher distribution
            $r_{g,t}$ using $(x^H,q,y_{<t}^{(g)})$
            \STATE Select the teacher-selected top-$K$ support
            $\mathcal{S}_{g,t}^{K}\leftarrow
            \operatorname{TopK}(r_{g,t})$
            \STATE Compute the bias-corrected teacher-top-$K$ reverse KL
            $d_K(p_{g,t}\|r_{g,t})$ using
            Eq.~\eqref{eq:corrected-rkl}
        \ENDFOR
    \ENDFOR
    \STATE Compute $\mathcal{L}_{\mathrm{RP\text{-}OPSD}}$
    using Eq.~\eqref{eq:rp-opsd-objective}
    \STATE Update the student:
    $\theta\leftarrow\theta-\eta\nabla_\theta
    \mathcal{L}_{\mathrm{RP\text{-}OPSD}}$
    \STATE Update the teacher:
    $\phi\leftarrow(1-\rho)\phi+\rho\theta$
    \STATE Synchronize the rollout policy:
    $\theta^-\leftarrow\theta$
\ENDFOR
\STATE \textbf{return} $\pi_\theta^L$
\end{algorithmic}
\end{algorithm}

\section{Experiments}
\label{sec:experiments}

\subsection{Experimental Setup}
\label{sec:experimental-setup}

\paragraph{Training Settings.}
We apply RP-OPSD to Qwen3.5-4B and Qwen3.5-9B~\cite{qwen2026qwen35}. The training set contains
5.2K samples drawn from Vision-SR1~\cite{li2026visionsr1},
VLM-CapCurriculum Perception~\cite{wu2026vlmcapcurriculum},
ZwZ-RL-VQA~\cite{wei2026zooming}, and
Vision-OPD~\cite{visionopd},see \textbf{Supplement}.Training Data for the detailed composition. The teacher receives the original image, whereas the student
receives an image whose width and height are each reduced by a factor of two
using Lanczos interpolation. All parameters, including the vision encoder,
are trainable. We optimize the bias-corrected teacher-top-100 reverse KL
objective and update the teacher using an exponential moving average (EMA)
with a rate of 0.05. For each input, we sample eight responses with a
temperature of 1.0, top-$p$ of 1.0, no top-$k$ truncation, and a maximum
generation length of 1,024 tokens. Each rollout batch is used for a single
student update, after which the rollout policy is synchronized with the
updated student. All models are trained for one epoch, corresponding to 55
optimization steps, with a batch size of 96 and a learning rate of
$2\times10^{-6}$. We apply linear warmup for the first 10 steps and set both
the data and generation seeds to 42. Training is conducted on 8$\times$ H20
GPUs.

\paragraph{Benchmarks.}
We consider two groups of benchmarks. The first group evaluates fine-grained
visual perception and includes V$^{*}$Bench~\cite{wu2023vstar},
HR-Bench 4K/8K~\cite{wang2024hrbench},
MME-RealWorld EN/CN~\cite{zhang2025mmerealworld}, and
VisualProbe~\cite{lai2025minio3}. The second group evaluates generalization
beyond the training distribution and includes MMVP~\cite{tong2024mmvp},
MMStar~\cite{chen2024mmstar}, and POPE~\cite{li2023pope}. The ablation
studies additionally report CV-Bench~\cite{tong2024cambrian}. All
models are evaluated using original-resolution images. We use rule-based
parsing whenever applicable; the remaining responses are assessed by a
Qwen3.5-9B~\cite{qwen2026qwen35} LLM judge using its default decoding settings.
\begin{table*}[t]
\centering
\caption{Main results on Qwen3.5-9B and Qwen3.5-4B. All methods are evaluated
with original-resolution images. Avg. is the unweighted mean of the nine
reported metrics. Bold denotes the best result within each model scale.}
\label{tab:main-results}
\begingroup
\small
\setlength{\tabcolsep}{2.8pt}
\begin{tabular*}{\textwidth}{@{\extracolsep{\fill}}lrrrrrrrrrr@{}}
\toprule
Model &
\multicolumn{1}{c}{V$^{*}$} &
\multicolumn{1}{c}{HR-4K} &
\multicolumn{1}{c}{HR-8K} &
\multicolumn{1}{c}{\shortstack{MME-RW\\EN}} &
\multicolumn{1}{c}{\shortstack{MME-RW\\CN}} &
\multicolumn{1}{c}{\shortstack{Visual\\Probe}} &
\multicolumn{1}{c}{MMVP} &
\multicolumn{1}{c}{MMStar} &
\multicolumn{1}{c}{POPE} &
\multicolumn{1}{c}{Avg.} \\
\midrule
\rowcolor[gray]{0.94}
\multicolumn{11}{c}{\textit{Qwen3.5-9B}} \\
Base
  & 84.82 & 84.75 & 81.50 & 71.40 & 67.67 & 41.85 & 83.00 & 82.07 & 89.36 & 76.27 \\
SFT
  & 91.10 & \textbf{87.88} & 83.62 & 73.25 & 71.54 & 51.25 & 83.67 & 78.93 & 89.74 & 79.00 \\
GRPO
  & 88.48 & 84.50 & 81.50 & 75.72 & 71.81 & 50.38 & \textbf{84.33} & 80.73 & 89.09 & 78.50 \\
OPSD
  & 91.10 & 86.88 & 84.25 & \textbf{78.12} & \textbf{74.43} & 49.97 & 83.00 & 81.40 & 89.08 & 79.80 \\
Vision-OPD
  & 85.86 & 86.62 & \textbf{85.12} & 73.40 & 70.46 & 56.84 & 81.33 & 81.53 & \textbf{89.79} & 78.99 \\
\textbf{RP-OPSD}
  & \textbf{91.10} & 86.50 & 84.12 & 76.91 & 72.84 & \textbf{56.97} & 83.33 & \textbf{82.67} & 89.43 & \textbf{80.43} \\
\midrule
\rowcolor[gray]{0.94}
\multicolumn{11}{c}{\textit{Qwen3.5-4B}} \\
Base
  & 84.29 & 84.38 & 80.13 & 64.20 & 63.80 & 43.22 & 76.67 & 78.53 & 88.28 & 73.72 \\
SFT
  & 87.96 & \textbf{85.75} & 80.00 & 71.24 & 69.17 & 48.53 & 78.00 & 68.60 & \textbf{89.43} & 75.41 \\
GRPO
  & 84.82 & 81.88 & 78.38 & 72.15 & 70.27 & 52.23 & \textbf{82.33} & 72.27 & 85.63 & 75.55 \\
OPSD
  & 86.39 & 85.50 & 80.12 & \textbf{76.75} & \textbf{74.72} & 48.78 & 82.00 & 76.67 & 88.70 & 77.74 \\
Vision-OPD
  & 87.96 & 82.62 & 81.75 & 74.70 & 70.46 & \textbf{54.95} & 77.00 & 77.40 & 89.10 & 77.33 \\
\textbf{RP-OPSD}
  & \textbf{87.96} & 85.50 & \textbf{82.00} & 76.56 & 72.42 & 54.37 & 78.33 & \textbf{79.07} & 88.98 & \textbf{78.35} \\
\bottomrule
\end{tabular*}
\endgroup
\end{table*}
\paragraph{Baselines.}
We compare RP-OPSD with \textbf{Base}, \textbf{SFT}, \textbf{GRPO},
\textbf{OPSD}, and \textbf{Vision-OPD}. Base denotes the original Qwen3.5
model without post-training and measures the overall gains from training.
SFT uses the original images and ground-truth answers and minimizes
cross-entropy over the assistant tokens. GRPO~\cite{shao2024deepseekmath} samples eight responses per
original-resolution image and computes group-relative advantages using a
binary correctness reward. OPSD~\cite{zhao2026opsd} presents the same original image to the
student and teacher while augmenting the teacher prompt with an answer hint
constructed from the ground truth. The teacher then provides token-level
distillation signals on the student's on-policy responses. SFT, GRPO, OPSD,
and RP-OPSD use the same 5.2K training samples. Vision-OPD~\cite{visionopd} performs
region-to-global self-distillation, where the teacher receives an evidence
crop and the student receives the full image with bounding-box annotations.
We reproduce Vision-OPD using its official implementation.

\subsection{Main Results}
\label{sec:main-results}

As shown in Table~\ref{tab:main-results}, RP-OPSD achieves the highest average
score at both model scales. With the 9B model, RP-OPSD reaches 80.43,
outperforming Base, Vision-OPD, and OPSD by 4.16, 1.44, and 0.63 points,
respectively. With the 4B model, the corresponding improvements are 4.63, 1.03,
and 0.61 points. These results indicate that the benefits of RP-OPSD generalize
across model scales.

Compared with Base, 9B RP-OPSD improves V$^{*}$Bench, VisualProbe,
MME-RW EN/CN, and HR-Bench 4K/8K by 6.28, 15.12, 5.51/5.17, and
1.75/2.62 points, respectively. The 4B model improves MME-RW EN/CN and
VisualProbe by 12.36, 8.62, and 11.15 points, respectively. Meanwhile, the
9B model improves MMVP, MMStar, and POPE by 0.33, 0.60, and 0.07 points,
suggesting that the gains in fine-grained perception do not come at the cost
of general visual capabilities. Relative to OPSD, RP-OPSD improves the average
score by 0.63 points and yields gains of 7.00 and 1.27 points on VisualProbe
and MMStar, respectively. This result indicates that resolution-privileged
distillation effectively enhances fine-grained visual search and general
visual perception.

\subsection{Low-Resolution Capability Analysis}
\label{sec:low-resolution-evaluation}

\begin{table}[t]
\centering
\caption{Performance of Qwen3.5-9B and RP-OPSD under half-resolution evaluation. Avg. is the unweighted
mean of the five reported metrics, and $\Delta$ denotes the improvement over
Base. Higher values are better for all metrics. Bold denotes the better score
in each row.}
\label{tab:half-resolution-evaluation}
\begingroup
\small
\setlength{\tabcolsep}{3pt}
\begin{tabular*}{\columnwidth}{@{\extracolsep{\fill}}lrrr@{}}
\toprule
Benchmark &
\multicolumn{1}{c}{\shortstack{Qwen3.5-9B}} &
\multicolumn{1}{c}{\shortstack{RP-OPSD}} &
\multicolumn{1}{c}{$\Delta$} \\
\midrule
VisualProbe
  & 34.81 & \textbf{49.71} & $+14.90$ \\
V$^{*}$Bench
  & 69.63 & \textbf{81.68} & $+12.05$ \\
HR-Bench 4K
  & 79.25 & \textbf{82.00} & $+2.75$ \\
POPE
  & 87.83 & \textbf{88.39} & $+0.56$ \\
CV-Bench
  & 85.75 & \textbf{85.94} & $+0.19$ \\
\midrule
\textbf{Avg.}
  & 71.45 & \textbf{77.54} & \textbf{$+6.09$} \\
\bottomrule
\end{tabular*}
\endgroup
\end{table}
As shown in Table~\ref{tab:half-resolution-evaluation}, RP-OPSD improves all
five benchmarks under low-resolution evaluation, increasing the average score
from 71.45 to 77.54. The largest gains occur on VisualProbe and
V$^{*}$Bench, where RP-OPSD improves the base model by 14.90 and 12.05
points, respectively. These benchmarks emphasize small visual details and
targeted visual search, indicating that supervision from the
original-resolution teacher enables the student to make better use of limited
visual evidence.

Together with the original-resolution results in
Table~\ref{tab:main-results}, these findings support our central hypothesis:
RP-OPSD directly strengthens the model under the low-resolution student view,
and the resulting capability improvement transfers to original-resolution inference. The gains therefore reflect an improvement in the model itself.

\subsection{Ablation Studies and Analysis}
\label{sec:ablation-analysis}

\subsubsection{Choice of Distillation Objective}
\label{sec:ablation-objective}

Table~\ref{tab:distillation-objective} compares GSD~\cite{agarwal2024gkd},
forward KL, standard reverse KL, and bias-corrected teacher-top-100 reverse KL
to examine how the token-level distillation objective and top-$K$
truncation-bias correction affect performance. We use Qwen3.5-9B, set
$\alpha=0.5$ for GSD, and keep all other training and evaluation settings
fixed.

\begin{table*}[t]
\centering
\caption{Ablation on the distillation objective. All objectives use the
teacher-selected Top-100 support; RKL denotes reverse KL. Avg. is the
unweighted mean of the eight metrics. Bold marks the best result in each
column; for ties involving the default setting, only the default is bolded.}
\label{tab:distillation-objective}
\begingroup
\small
\setlength{\tabcolsep}{3pt}
\begin{tabular*}{\textwidth}{@{\extracolsep{\fill}}p{3.6cm}rrrrrrrrr@{}}
\toprule
Setting &
\multicolumn{1}{c}{V$^{*}$} &
\multicolumn{1}{c}{HR-4K} &
\multicolumn{1}{c}{HR-8K} &
\multicolumn{1}{c}{\shortstack{Visual\\Probe}} &
\multicolumn{1}{c}{MMVP} &
\multicolumn{1}{c}{CV-Bench} &
\multicolumn{1}{c}{MMStar} &
\multicolumn{1}{c}{POPE} &
\multicolumn{1}{c}{Avg.} \\
\midrule
GSD Top-100
  & 91.10 & 86.38 & 82.38 & 55.89 & 83.33 & \textbf{87.91} & 82.47 & 89.42 & 82.36 \\
Forward Top-100 KL
  & 90.05 & 85.62 & 81.88 & 53.84 & 82.33 & 87.61 & \textbf{83.40} & 89.39 & 81.77 \\
Reverse Top-100 KL
  & 89.01 & \textbf{87.25} & 82.75 & 53.79 & 82.33 & 87.86 & 80.80 & 88.93 & 81.59 \\
Bias-Corrected RKL (Ours)
  & \textbf{91.10} & 86.50 & \textbf{84.12} & \textbf{56.97} & \textbf{83.33} & 87.73 & 82.67 & \textbf{89.43} & \textbf{82.73} \\
\bottomrule
\end{tabular*}
\endgroup
\end{table*}

The bias-corrected teacher-top-100 reverse KL objective achieves the highest
average score of 82.73, outperforming standard reverse KL and GSD by 1.14 and
0.37 points, respectively. Relative to standard reverse KL, it improves
V$^{*}$Bench, HR-Bench 8K, and VisualProbe by 2.09, 1.37, and 3.18 points.
These results support correcting top-$K$ truncation bias when distillation is
restricted to the teacher-selected vocabulary support. We therefore adopt
this objective in subsequent experiments.

\subsubsection{Teacher Update Strategy}
\label{sec:teacher-update}

Table~\ref{tab:teacher-update} compares an EMA teacher with a teacher fixed at
its initial parameters to examine whether EMA provides a more effective
distillation target. Both variants use the same initialization, visual inputs,
distillation objective, and training budget.

\begin{table*}[t]
\centering
\caption{Ablation on the teacher update strategy. Avg. is the unweighted mean
of the eight metrics. Bold marks the best result in each column; for ties
involving the default setting, only the default is bolded.}
\label{tab:teacher-update}
\begingroup
\small
\setlength{\tabcolsep}{3pt}
\begin{tabular*}{\textwidth}{@{\extracolsep{\fill}}p{3.6cm}rrrrrrrrr@{}}
\toprule
Setting &
\multicolumn{1}{c}{V$^{*}$} &
\multicolumn{1}{c}{HR-4K} &
\multicolumn{1}{c}{HR-8K} &
\multicolumn{1}{c}{\shortstack{Visual\\Probe}} &
\multicolumn{1}{c}{MMVP} &
\multicolumn{1}{c}{CV-Bench} &
\multicolumn{1}{c}{MMStar} &
\multicolumn{1}{c}{POPE} &
\multicolumn{1}{c}{Avg.} \\
\midrule
Frozen Initial Teacher
  & \textbf{91.62} & 86.50 & 83.00 & 52.39 & \textbf{84.33} & \textbf{88.23} & 82.00 & \textbf{89.69} & 82.22 \\
EMA Teacher (Ours)
  & 91.10 & \textbf{86.50} & \textbf{84.12} & \textbf{56.97} & 83.33 & 87.73 & \textbf{82.67} & 89.43 & \textbf{82.73} \\
\bottomrule
\end{tabular*}
\endgroup
\end{table*}

\begin{table*}[!t]
\centering
\caption{Ablation on the student training resolution. The teacher and
evaluation use original-resolution images. Avg. is the unweighted mean of the
eight metrics. Bold marks the best result in each column; for ties involving
the default setting, only the default is bolded.}
\label{tab:student-resolution}
\begingroup
\small
\setlength{\tabcolsep}{3pt}
\begin{tabular*}{\textwidth}{@{\extracolsep{\fill}}p{3.6cm}rrrrrrrrr@{}}
\toprule
Setting &
\multicolumn{1}{c}{V$^{*}$} &
\multicolumn{1}{c}{HR-4K} &
\multicolumn{1}{c}{HR-8K} &
\multicolumn{1}{c}{\shortstack{Visual\\Probe}} &
\multicolumn{1}{c}{MMVP} &
\multicolumn{1}{c}{CV-Bench} &
\multicolumn{1}{c}{MMStar} &
\multicolumn{1}{c}{POPE} &
\multicolumn{1}{c}{Avg.} \\
\midrule
$1/3$ width/height
  & \textbf{91.62} & 86.75 & 83.12 & 55.38 & 82.67 & 86.55 & 80.93 & 88.40 & 81.93 \\
$1/4$ width/height
  & 89.01 & 85.12 & 82.75 & 55.21 & 79.00 & 85.85 & 79.87 & 84.57 & 80.17 \\
$1/2\rightarrow1$ canvas
  & 91.10 & \textbf{86.88} & 83.25 & 51.82 & 82.00 & 87.66 & 82.27 & \textbf{89.82} & 81.85 \\
$1/2$ width/height (Ours)
  & 91.10 & 86.50 & \textbf{84.12} & \textbf{56.97} & \textbf{83.33} & \textbf{87.73} & \textbf{82.67} & 89.43 & \textbf{82.73} \\
\bottomrule
\end{tabular*}
\endgroup
\end{table*}

\begin{table}[!t]
\centering
\caption{Training efficiency under identical data, batch size (96), rollouts
per input (8), and hardware (8$\times$ NVIDIA H20 GPUs). Speedup is
$T_{\mathrm{OPSD}}/T_{\mathrm{method}}$; Full and Half denote input resolution.}
\label{tab:training-efficiency}
\begingroup
\small
\setlength{\tabcolsep}{2pt}
\begin{tabular*}{\columnwidth}{@{\extracolsep{\fill}}lccrrr@{}}
\toprule
Method &
Student &
Teacher &
\multicolumn{1}{c}{Time (h)} &
\multicolumn{1}{c}{Speedup} &
\multicolumn{1}{c}{\shortstack{Main\\Avg.}} \\
\midrule
OPSD
  & Full
  & Full+hint
  & 13.93
  & 1.00$\times$
  & 79.80 \\
\textbf{RP-OPSD}
  & Half
  & Full
  & \textbf{7.83}
  & \textbf{1.78$\times$}
  & \textbf{80.43} \\
\bottomrule
\end{tabular*}
\endgroup
\end{table}

The EMA teacher reaches 82.73 on average, 0.51 points above the frozen
teacher, with gains of 1.12, 4.58, and 0.67 on HR-Bench 8K, VisualProbe,
and MMStar. Although the frozen teacher is slightly better on V$^{*}$Bench,
MMVP, and CV-Bench, EMA provides stronger average and fine-grained performance.

\subsubsection{Student Input Resolution}
\label{sec:student-resolution}

Table~\ref{tab:student-resolution} varies only the student resolution; the
teacher and evaluation retain original-resolution images. The
$1/2\rightarrow1$ setting downsamples both dimensions by half and then
restores the original canvas size, allowing us to separate lost image detail
from a change in the input canvas.

Halving both dimensions yields the best average score of 82.73. Reducing them
by factors of three and four lowers the average by 0.80 and 2.56 points,
showing that aggressive downsampling weakens the visual evidence available for
transfer. The $1/2\rightarrow1$ setting reaches only 81.85, below direct
half-resolution training, confirming that restoring the canvas does not
recover lost visual details.

\subsection{Training Efficiency}
\label{sec:training-efficiency}

Table~\ref{tab:training-efficiency} compares OPSD and RP-OPSD under identical
data, steps, batch size, rollouts, and hardware, isolating the effect of the
student input resolution.

Let $n_H$ be the original-resolution visual-token count; halving both image
dimensions gives $n_L=n_H/4$. Let $C_g(n)$ and $C_f(n)$ denote one rollout
and one full-sequence forward pass, respectively. With $G$ rollouts and a
$3C_f(n)$ forward--backward approximation, OPSD costs
\begin{equation}
\mathcal{F}_{\mathrm{OPSD}}\simeq G\!\left[C_g(n_H)+4C_f(n_H)\right].
\label{eq:opsd-flops}
\end{equation}
whereas RP-OPSD costs
\begin{equation}
\mathcal{F}_{\mathrm{RP}}\simeq G\!\left[C_g(n_H/4)+3C_f(n_H/4)+C_f(n_H)\right].
\label{eq:rp-opsd-flops}
\end{equation}
Let $\Delta C_g=C_g(n_H)-C_g(n_H/4)$ and define $\Delta C_f$
analogously. Their gap is
\begin{equation}
\mathcal{F}_{\mathrm{OPSD}}-\mathcal{F}_{\mathrm{RP}}
\simeq G\!\left(\Delta C_g+3\Delta C_f\right)>0.
\label{eq:flops-reduction}
\end{equation}
The factor three approximates student forward--backward computation. Since
$n_L=n_H/4$, linear vision terms fall to one quarter and quadratic
self-attention terms to one sixteenth.Since text tokens are much fewer than visual tokens in our setting, we omit their contribution from the FLOPs estimate. The shared original-resolution teacher cancels in
Eq.~\eqref{eq:flops-reduction}, leaving only rollout and student-update
savings. The estimate holds nonvisual teacher context fixed and omits system
overheads, which the wall-clock measurement captures.

RP-OPSD reduces training time from 13.93 to 7.83 hours ($1.78\times$) while
raising the main-table average from 79.80 to 80.43. Thus, lower-resolution
student inputs reduce training cost without sacrificing original-resolution
performance. The reported speedup covers training only and excludes
evaluation.
\section{Conclusion}
We introduced RP-OPSD, an on-policy self-distillation framework that uses the
capability gap induced by original- and low-resolution views of the same image
as privileged supervision for multimodal large language models. A
low-resolution student generates on-policy trajectories, while an
original-resolution EMA teacher provides token-level distribution targets over
the same prefixes using richer visual evidence. RP-OPSD requires only
image--query pairs, without additional answer annotations, external teacher
models, generated solution traces, or localized visual evidence. Across
Qwen3.5-4B and Qwen3.5-9B, RP-OPSD achieves the best average performance among
the compared post-training methods, with relative improvements of 6.28\% and
5.45\% over their base models, respectively. Gains under half-resolution
evaluation further support the transfer of improved low-resolution capability
to original-resolution inference. RP-OPSD also yields a $1.78\times$ training
speedup over OPSD, showing that resolution differences provide simple,
effective, and efficient privileged information.
\bibliography{references}

\end{document}


\maketitle

\input{appendix/content}

%% file: appendix/content.tex
\appendix

\section{Training Data}
\label{app:training-data}

We construct the 5,295-example single-image training set from
Vision-SR1, VLM-CapCurriculum Perception, ZwZ-RL-VQA, and the public
Vision-OPD training set. Table~\ref{tab:appendix-data-composition}
reports the source composition.

\begin{table}[!h]
\centering
\small
\begin{tabular}{@{}lr@{}}
\toprule
Source & Samples \\
\midrule
ZwZ-RL-VQA & 3,034 \\
Vision-OPD & 1,060 \\
Vision-SR1 & 887 \\
VLM-CapCurriculum Perception & 314 \\
\midrule
\textbf{Total} & \textbf{5,295} \\
\bottomrule
\end{tabular}
\caption{Source composition of the RP-OPSD training set.}
\label{tab:appendix-data-composition}
\end{table}
\FloatBarrier

\section{Detailed Training Settings}
\label{app:training-settings}

We initialize the student and teacher from the same Qwen3.5 checkpoint.
The student generates on-policy responses from the half-resolution image.
The teacher then scores the same response prefixes while observing the
original-resolution image and the same text prompt. Gradients propagate
only through the student. After each actor update, the teacher parameters
are updated as
\begin{equation}
    \phi_{s+1}
    =
    (1-\rho)\phi_s+\rho\theta_{s+1},
    \qquad \rho=0.05.
\end{equation}
All student parameters, including the vision encoder, remain trainable.

We distill the teacher-selected Top-100 token distribution with the
bias-corrected reverse KL objective described in the main paper. For each
input, the student samples eight responses. Training uses no sequence-level
reward and no external judge. Tables~\ref{tab:appendix-model-config}
and~\ref{tab:appendix-optimization-config} list the complete configuration
shared by the 4B and 9B experiments.

We train with FSDP and vLLM under Python~3.12, PyTorch~2.10.0,
Transformers~5.5.0, vLLM~0.18.0, and Ray~2.53.0. Rollout uses eight
workers with tensor parallelism 1. The perception chat template prepends
an empty \texttt{<think></think>} block; it does not request or supervise
a reasoning trace. The 9B run requires 7.83 hours on eight H20 GPUs.

\begin{table}[!ht]
\centering
\small
\begin{tabular}{@{}ll@{}}
\toprule
Configuration & Value \\
\midrule
Backbone & Qwen3.5-4B / Qwen3.5-9B \\
Trainable parameters & All, including vision encoder \\
Student image & Physical $1/2$ width and height \\
Teacher image & Original resolution \\
Interpolation & Lanczos \\
Teacher update & EMA, $\rho=0.05$ \\
Distillation objective & Bias-corrected reverse KL \\
Distillation support & Teacher-selected Top-100 \\
Rollouts per input & 8 \\
Maximum response length & 1,024 tokens \\
Sampling temperature & 1.0 \\
Top-$p$ / Top-$k$ & $1.0$ / $-1$ \\
Maximum prompt length & 8,192 tokens \\
Maximum sequence length & 9,216 tokens \\
\bottomrule
\end{tabular}
\caption{Model and generation configuration shared by the 4B and 9B
experiments.}
\label{tab:appendix-model-config}
\end{table}

\begin{table}[!ht]
\centering
\small
\begin{tabular}{@{}ll@{}}
\toprule
Configuration & Value \\
\midrule
Global batch size & 96 \\
PPO mini-batch size & 96 \\
Micro-batch per GPU & 1 \\
Dynamic batching & Disabled \\
Optimizer & AdamW \\
Learning rate & $2\times10^{-6}$ \\
Adam coefficients & $(0.9, 0.999)$ \\
Weight decay & 0.01 \\
Gradient clipping & 1.0 \\
Linear warmup & 10 steps \\
Training schedule & 1 epoch / 55 steps \\
Checkpoint & Final step \\
Training precision & BF16 \\
Gradient checkpointing & Enabled \\
Actor parameter offload & Enabled \\
Optimizer offload & Enabled \\
Data / generation seed & 42 / 42 \\
Hardware & 8$\times$ NVIDIA H20 \\
\bottomrule
\end{tabular}
\caption{Optimization and system settings for RP-OPSD.}
\label{tab:appendix-optimization-config}
\end{table}
\FloatBarrier

\section{Case Study}
\label{app:case-study}

We compare Qwen3.5-9B Base and RP-OPSD on two original-resolution
VisualProbe cases.

\begin{figure*}[!t]
\centering
\begin{tabularx}{\textwidth}
{@{}p{0.62\textwidth}@{\hspace{0.03\textwidth}}X@{}}
\begin{minipage}[t]{\linewidth}
\vspace{0pt}
\centering
\includegraphics[
  width=\linewidth,
  height=0.34\textheight,
  keepaspectratio
]{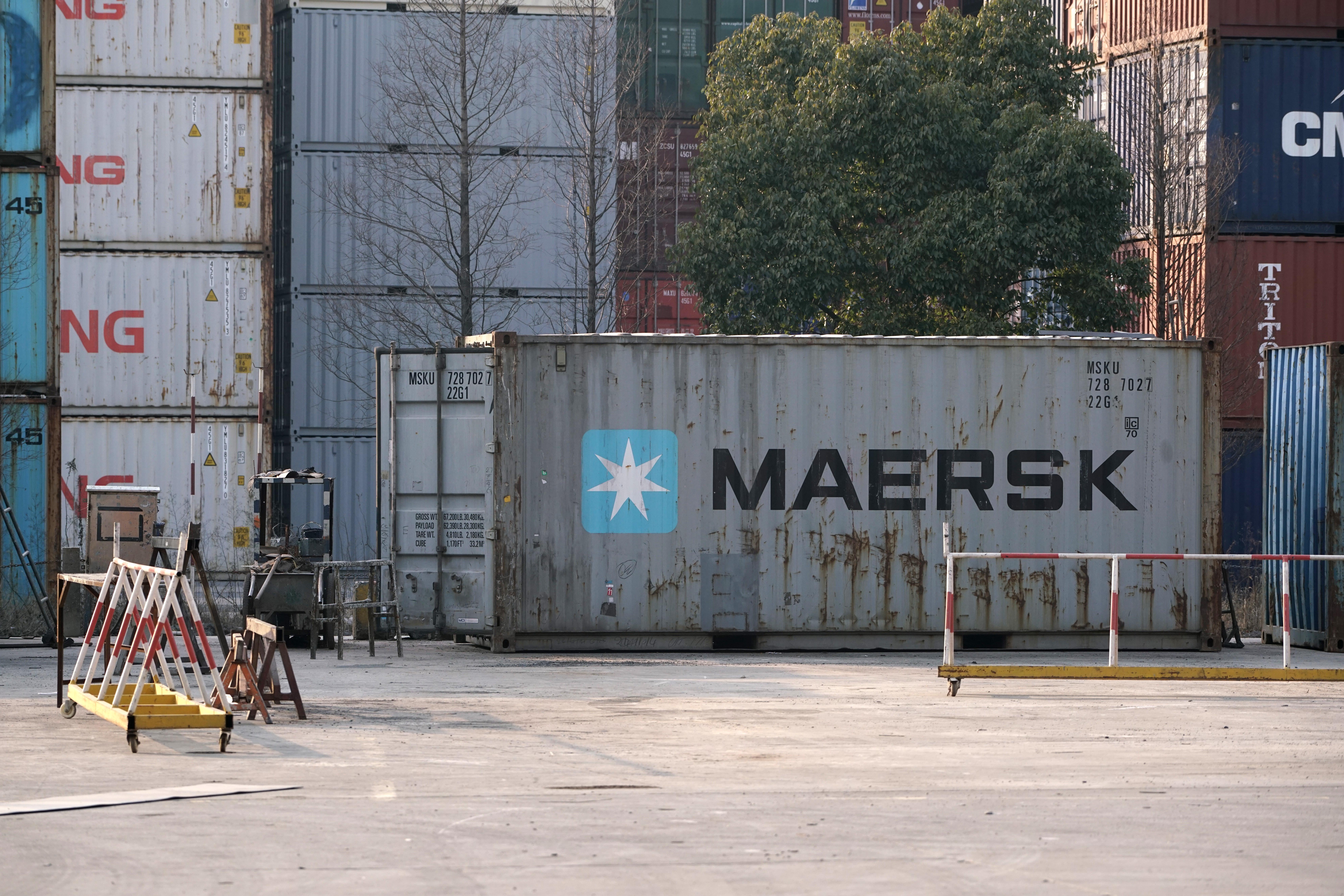}

\smallskip
\footnotesize (a) Full image, $7952\times5304$
\end{minipage}
&
\begin{minipage}[t]{\linewidth}
\vspace{0pt}
\centering
\includegraphics[
  width=\linewidth,
  height=0.22\textheight,
  keepaspectratio
]{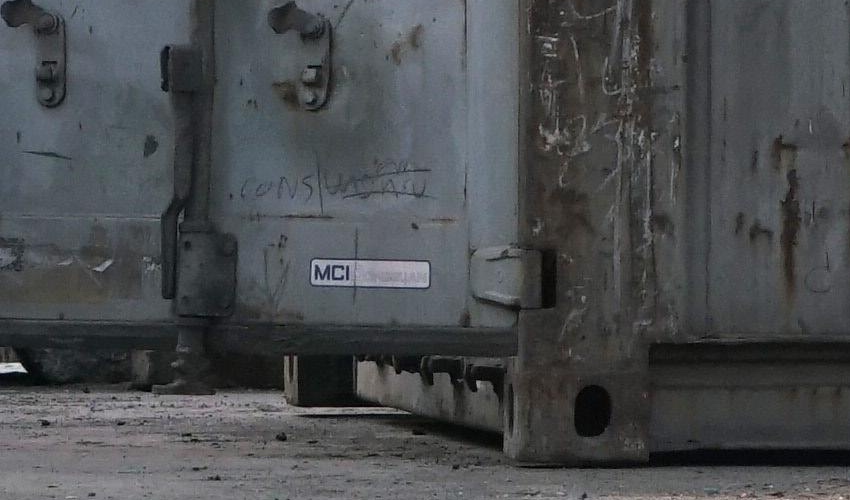}

\smallskip
\footnotesize (b) Target-region crop

\medskip
\raggedright
\small
\textbf{Question.}
What is written on the white rectangle at the bottom of the open door of
the shipping container?

\smallskip
\textbf{Ground truth.} \texttt{MCI}
\end{minipage}
\end{tabularx}

\medskip
\small
\begin{tabularx}{\textwidth}
{@{}>{\raggedright\arraybackslash}X
@{\hspace{0.04\textwidth}}
>{\raggedright\arraybackslash}X@{}}
\toprule
\textbf{Qwen3.5-9B Base (incorrect)}
&
\textbf{RP-OPSD (correct)}
\\
\midrule
\textbf{Prediction.} \texttt{MSKU 728 702 7 / 22G1}

\smallskip
\textbf{Analysis.} The base model copies the prominent container code
near the queried region instead of reading the small white label.
&
\textbf{Prediction.} \texttt{MCI}

\smallskip
\textbf{Analysis.} RP-OPSD localizes the white rectangle near the bottom
of the open door and transcribes the small target text correctly.
\\
\bottomrule
\end{tabularx}
\caption{Fine-grained OCR on a shipping container. The target crop makes
the localization requirement explicit, while the prediction comparison
shows that RP-OPSD resolves the small queried label.}
\label{fig:case-study-ocr}
\end{figure*}

\begin{figure*}[!t]
\centering
\begin{tabularx}{\textwidth}
{@{}p{0.62\textwidth}@{\hspace{0.03\textwidth}}X@{}}
\begin{minipage}[t]{\linewidth}
\vspace{0pt}
\centering
\includegraphics[
  width=\linewidth,
  height=0.34\textheight,
  keepaspectratio
]{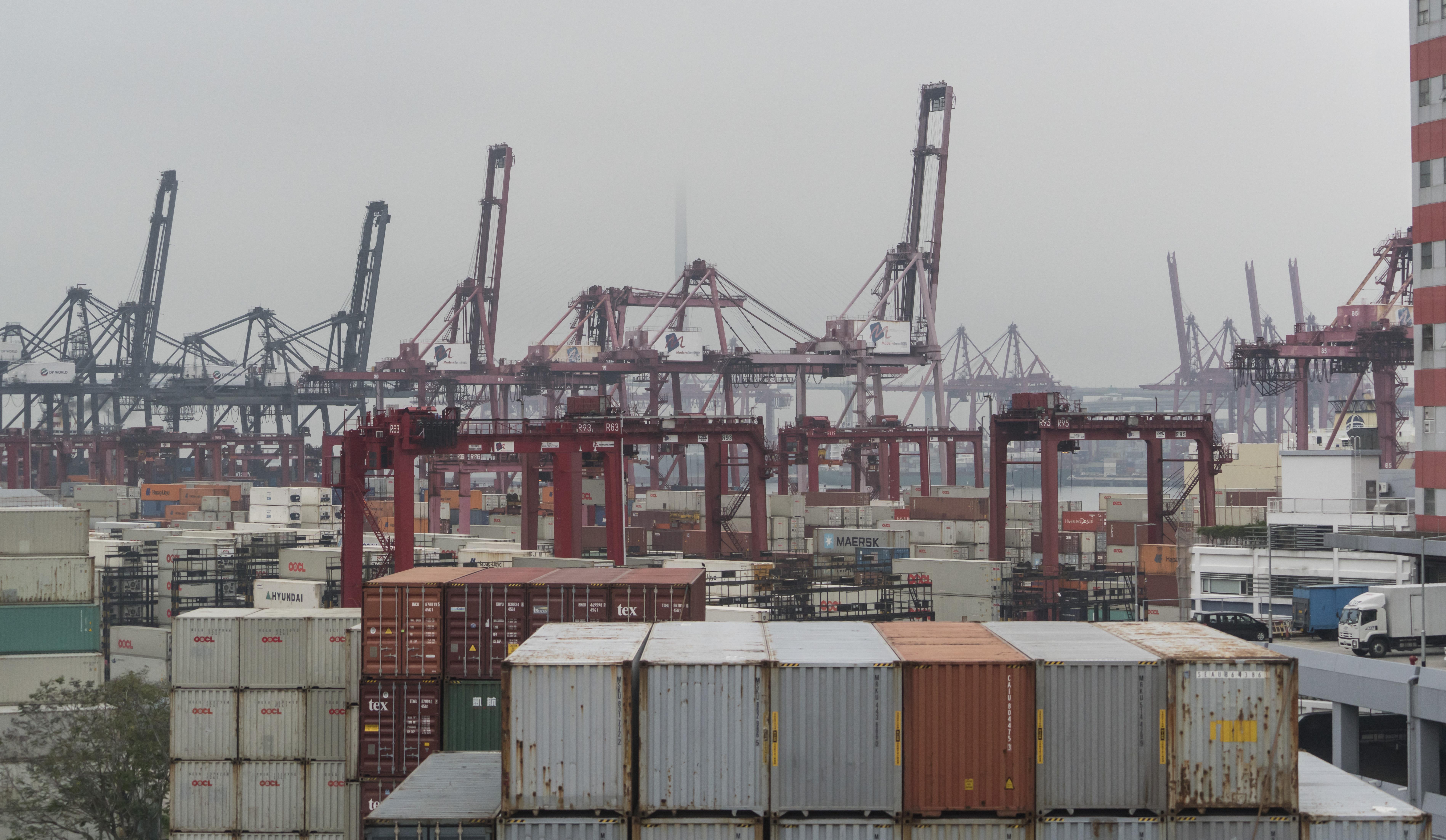}

\smallskip
\footnotesize (a) Full image, $7587\times4410$
\end{minipage}
&
\begin{minipage}[t]{\linewidth}
\vspace{0pt}
\centering
\includegraphics[
  width=\linewidth,
  height=0.22\textheight,
  keepaspectratio
]{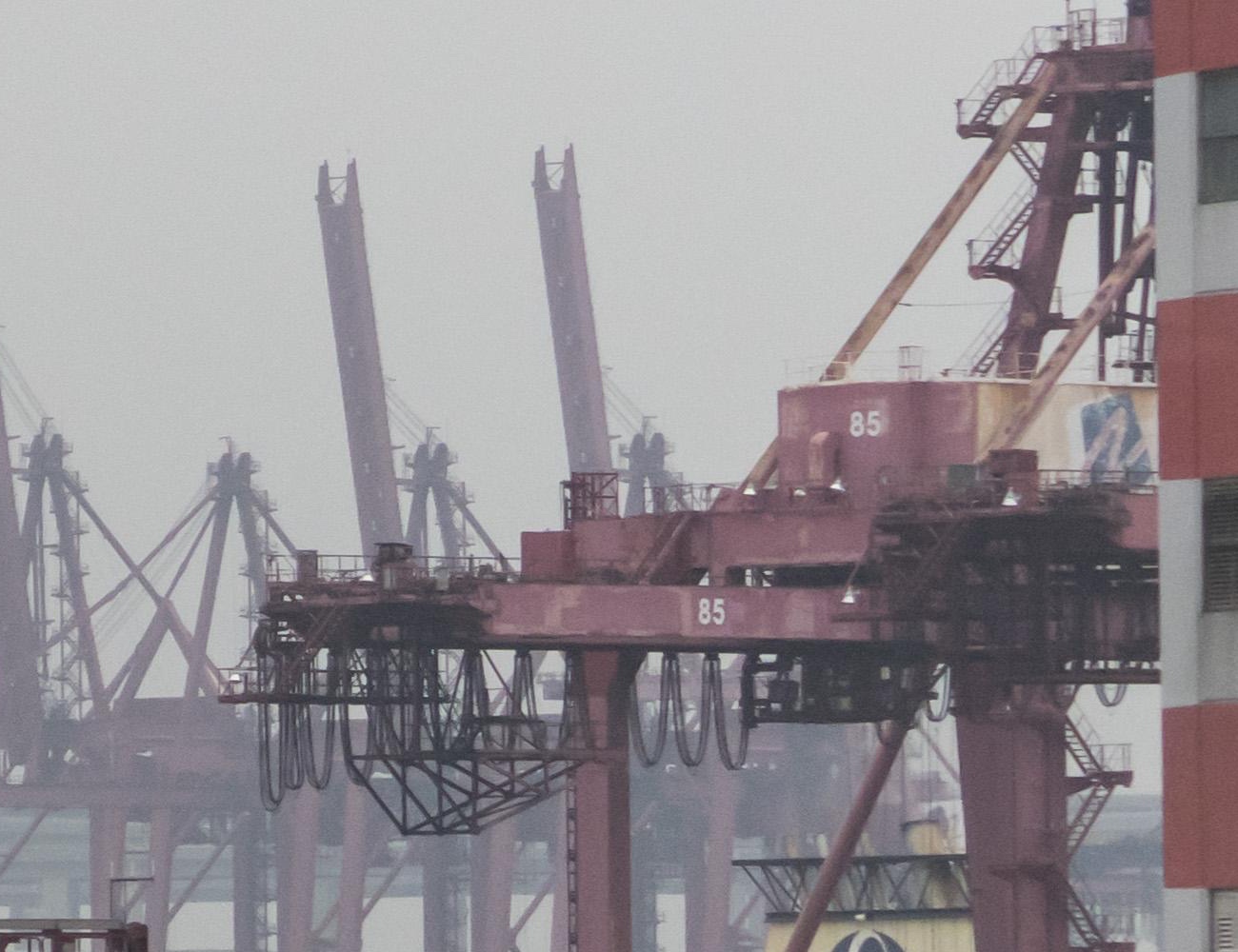}

\smallskip
\footnotesize (b) Target-region crop

\medskip
\raggedright
\small
\textbf{Question.}
What is the number on the red container crane on the far right?

\smallskip
\textbf{Ground truth.} \texttt{85}
\end{minipage}
\end{tabularx}

\medskip
\small
\begin{tabularx}{\textwidth}
{@{}>{\raggedright\arraybackslash}X
@{\hspace{0.04\textwidth}}
>{\raggedright\arraybackslash}X@{}}
\toprule
\textbf{Qwen3.5-9B Base (incorrect)}
&
\textbf{RP-OPSD (correct)}
\\
\midrule
\textbf{Prediction.} \texttt{R95}

\smallskip
\textbf{Analysis.} The base model selects a nearby crane identifier,
failing to ground the phrase ``on the far right'' to the intended crane.
&
\textbf{Prediction.} \texttt{85}

\smallskip
\textbf{Analysis.} RP-OPSD identifies the far-right crane and recognizes
the small white number on its upper structure.
\\
\bottomrule
\end{tabularx}
\caption{Distant number recognition in a container terminal. RP-OPSD
grounds the spatial reference to the correct crane and reads its small
identifier, whereas the base model selects a nearby label.}
\label{fig:case-study-crane}
\end{figure*}